\documentclass[conference]{IEEEtran}
\IEEEoverridecommandlockouts
\usepackage{cite}
\usepackage{amsmath,amssymb,amsfonts}
\usepackage{algorithmic}
\usepackage{graphicx}
\usepackage{textcomp}
\usepackage{xcolor}
\usepackage{tabularx}
\usepackage{booktabs}
\usepackage{nohyperref}
\usepackage{makecell} 
\usepackage{flushend} 
\usepackage{url}

\usepackage{array}

\def\BibTeX{{\rm B\kern-.05em{\sc i\kern-.025em b}\kern-.08em
    T\kern-.1667em\lower.7ex\hbox{E}\kern-.125emX}}
\begin{document}


\title{BERT-Assisted Semantic Annotation Correction for Emotion-Related Questions
}

\author{\IEEEauthorblockN{Abe Kazemzadeh}
  \IEEEauthorblockA{
    \textit{University of St. Thomas} \textit{Graduate Programs in Software} \\
St. Paul, MN U.S.A. \\
abe.kazemzadeh@stthomas.edu, orcid:0000-0002-1851-294X}
}


\maketitle

\begin{abstract}

Annotated data have traditionally been used to provide the input for
training a supervised machine learning (ML) model. However, current
pre-trained ML models for natural language processing
(NLP) contain embedded linguistic information that can be used to
inform the annotation process.
We use the BERT neural language model to feed information back into an
annotation task that involves semantic labelling of dialog behavior in
a question-asking game called Emotion Twenty Questions (EMO20Q).
First we describe the background of BERT, the EMO20Q data, and
assisted annotation tasks.  Then we describe the methods for
fine-tuning BERT for the purpose of checking the annotated labels. To
do this, we use the paraphrase task as a way to check that all
utterances with the same annotation label are classified as
paraphrases of each other.  We show this method to be an effective way to
assess and revise annotations of textual user data with complex,
utterance-level semantic labels.

\end{abstract}


\section{Introduction}

When annotating data for ML tasks, is not ideal to use only a single
annotator, especially for annotation tasks that involve complex or
subjective data. However, for reasons of expediency or cost, sometimes
using a single annotator is inevitable. This paper looks at an
instance of such cases. The basic contribution of this paper is to use
the predictions of a ML model on a test set as a way to assist
annotation by correcting annotation errors.

When an annotated test set is used to evaluate a trained model, the
errors will be either due to the ML model's performance or the
annotator's performance. In less performant models, more error is due
to the model, while in more performant models, more error can be
attributed to the annotator(s).

In scenarios where there are multiple annotators, one can assess and
model the agreement among annotators \cite{Mower2009,Davani2021},
which minimizes the effect of annotator uncertainty, inconsistency,
and other sources of errors like data or label ambiguity. In these
cases, we can be reasonably certain about the ground truth of the
annotation or at least the margin of inter-annotator agreement. In
these cases, error in the performance of the model is due to the model
itself (including being trained with insufficient data) or the
limitations of the task (e.g. task ambiguity or lack of annotator
training).

In single annotator cases, there are usually constraints to the amount
of annotations available due to the limitations of a single
individual's time and effort.
Recently though, using small
data sets to train modern ML models has become feasible and performant
by using fine-tuning.  This paper demonstrates the high performance of
modern ML models with the BERT neural ML model.  However, we focus on
an orthogonal issue, how the model predictions can inform the
annotation task.


The inspiration for looking at the use of machine learning to assist
the annotation task comes from observing the remarkable performance of
modern ML models.  Applying BERT to our single-annotator project
showed 99.97\% accuracy (98.65\% precision and 98.46\% recall).  Upon
inspection of the errors, some of the errors turned out to be
annotation and anonymization errors, not classification errors.  In
essence, the machine-learned model helped to correct human annotation
errors and faulty data preprocessing errors.

Our main motivation for using  machine learned models
to assist the annotation process is that current state-of-the-art
neural language models have demonstrated near human performance on a
number of tasks \cite{Firestone2020,Eckersley2017}.  If human-level
performance is reached, the remaining error in learning from annotations
should be due to ambiguity in the annotation task and inconsistency in
performing the task, i.e. human error.

Another reason for attempting  machine assistance of annotation is for
difficult annotation tasks.  The task we describe in this
paper has 700 unique sentence-level annotations.  The annotations are
an expressive domain language for the data they describe.  Moreover,
new data can be uniquely different than existing data, which can
necessitate creating new annotation labels. Annotations taks like this
can benefit from standardization from automated assistance.

One final reason for considering machine assistance of annotation is
for tasks where it is difficult to find annotators.  Researchers of
resource-limited languages and applications developed by single
researchers often do not have the ability to recruit annotators to
establish agreement and coverage.  Machine-assisted annotation
promises a way to bootstrap annotation of low resource languages and
enable smaller teams to make progress on new tasks.

In this paper we describe this experiment that helped correct
annotations errors in our data in Sections~\ref{sec:methodology} and
\ref{sec:results}.  Background information about BERT, EMO20Q, and
complex annotation tasks is given in
Sections~\ref{sec:bert-background}, \ref{sec:emo20q-background}, and
\ref{sec:annotation-background}, respectively.

\section{Background}
\subsection{BERT}
\label{sec:bert-background}

Recent  ML models for natural language processing
\cite{Vaswani2017} rely on pre-training, which takes large quantities
of text to form a language model.  The data used for pre-training
simply contain language information, not labels.  There is no
annotation involved in this pre-training process, but the process is
usually called \emph{self-supervised} rather than \emph{unsupervised}
because the language data itself is the data used for training.  The
sequence of words themselves provide the supervision through masking
and next sentence prediction tasks.  In the masking task, words are
randomly excluded and then predicted.  In the next sentence task, one
sentence is given and the next sentence in a document is predicted.
In this way the resultant ML model can be termed self-supervised
rather than unsupervised.

Due to the self-supervised nature of neural language models like BERT,
these model come to learn general information about the language that
can be leveraged in subsequent supervised \emph{fine-tuning} steps.
The language model training procedures of masking and next sentence
prediction aim to capture word occurrence statistics. However, the
vector representations and attention components that underlie these
models contain latent syntactic and semantic information
\cite{Manning2020}.  It is this latent information that enables these
models to obtain state of the art performance on many tasks.

While the self-supervised pre-training phase of building these neural
language models does not include annotations, the next step,
fine-tuning, usually does.  In the fine-tuning step, ML
models like classifiers or taggers can be trained from data and
annotation labels.  Traditional ML models would take the
input data and labels to learn a model.  Newer neural deep learning
models use the pre-trained models \cite{Turc2019} to encode the input,
converting the input data to a vector.  This encoding step embeds much
information about the target data and allows a larger, unannotated
dataset used during the pre-training step to inform the supervised
fine-tuning step. Although the traditional and modern approaches
differ, in both the traditional approach (data + annotation
\(\rightarrow\) model) and the modern approach ((data \(\rightarrow\)
encoder) + annotations \(\rightarrow\) model), information flows from
data and annotations to the machine-learned models.

\begin{figure*}[!t]
    \centering
    \includegraphics[width=.75\textwidth]{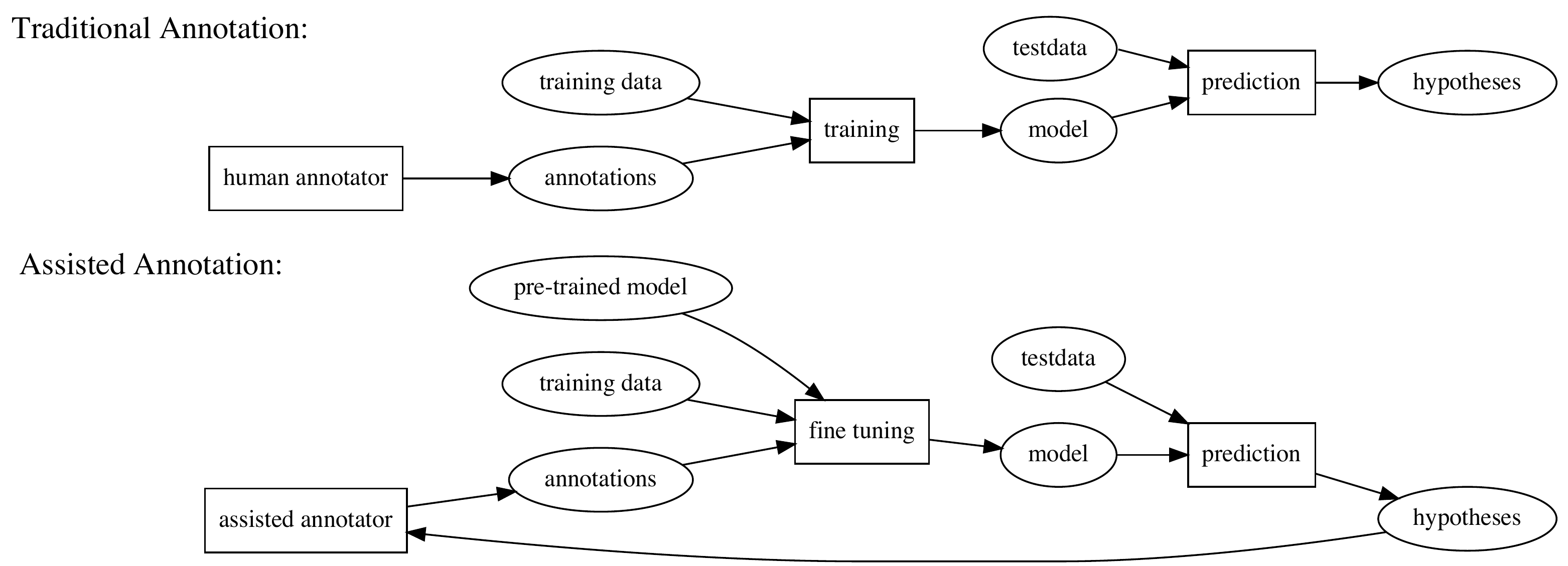}
    \caption{Flow chart illustrating the traditional approach, where
      annotations provide information used for ML training (top), and
      the proposed approach, where the ML model provides information
      back to the annotation process.}
    \label{annotation_graphviz}
\end{figure*}

In this paper we propose a method for reversing this flow of
information, so that information from the model's predictions flows
back to the annotator, as illustrated in
Fig.~\ref{annotation_graphviz}.  In this proposed method, we do train
a model from annotations, but we use the trained model to check and
re-label the input data.  Thus, instead of having a directed acyclic
graph (DAG) proceeding from annotation to model training, we have a
cycle where the annotations can be revised due to the model output.

\subsection{EMO20Q}
\label{sec:emo20q-background}

We consider an annotation task from the author's prior work: designing
a dialog agent to play the Emotion Twenty Questions game (EMO20Q)
\cite{Kazemzadeh2013a}. EMO20Q is a game like twenty questions where
the players are limited to emotion words.  The game is a vehicle to
collect natural language descriptions of emotion, which are then used
to train a dialog agent to play EMO20Q.  The transcripts from the
games provide rich descriptions of human emotions, behaviors, and
social situations. The data collected from 25 participants in
human-human EMO20Q games were used to create artificial dialog agents
that play the game and in turn acquire more data. In the
human-computer games, the 116 human participants picked the emotion
words and the computer asked the questions. An example of the game is
shown in Table~\ref{tab:exampledialog}.

\begin{table}[tb]
  \centering
  \caption{Example dialog from the EMO20Q questioner agent. The human
    interlocutor has chosen ``relaxation''
    \cite{Kazemzadeh2013a}.\label{tab:exampledialog}}

\begin{tabular}{|c|l|}
\hline 
Speaker/Event & Utterance/Content\tabularnewline
\hline 
Agent & Let me know when you are ready...\tabularnewline
User & ok\tabularnewline
Agent & Is it very intense?\tabularnewline
User & no\tabularnewline
Agent & is it positive?\tabularnewline
User & yes\tabularnewline
Agent & is it an emotion that just pops up spontaneously \tabularnewline
 & (vs being triggered by something)? \tabularnewline
User & depends\tabularnewline
Agent & is it caused by something in particular?\tabularnewline
User & well, yes, you do need the right conditions\tabularnewline
Agent & Is it relaxation?\tabularnewline
User & Yes\tabularnewline
Agent & so did I get it right?\tabularnewline
User & Yes, you got it\tabularnewline
Agent & Awesome! Would you like to play again?\tabularnewline
User & No, thanks.\tabularnewline
Agent & Thank you for playing\tabularnewline
\hline 
\end{tabular}

\end{table}

\begin{table}
  \caption{Example semantic annotations of EMO20Q questions
    \cite{Kazemzadeh2013a}. }

  {
    \centering
\begin{tabular}{|l p{.285\textwidth}|} 
\hline  
{\bf Semantic Annotation} & {\bf Examples} 
\tabularnewline 

cause(emptySet,e) & 
\emph{can you feel the emotion without any external events that cause it?} \newline
\emph{is it an emotion that just pops up spontaneously (vs being triggered by something)?} 
\tabularnewline 

cause(otherPerson,e) & 
\emph{is it caused by the person that it's directed at?} \newline
\emph{Do you need someone to pull this emotion out of you or evoke it? if so, who is it?}
\tabularnewline 

e.valence==negative & 
\emph{is it considered a negative thing to feel?} \newline
\emph{2) is it a negative emotion?}
\tabularnewline 

situation(e,birthday) & 
\emph{would you feel this if it was your birthday?} \newline
\emph{is it a socially acceptable emotion, say, at a birthday party?}
\tabularnewline 

e==frustration &
\emph{oh, is it frustrated?} \newline
\emph{frustration?}
\tabularnewline 
\hline 
\end{tabular}
}
  \label{tab:annotationexample}
\end{table}

Example annotations of the EMO20Q data is shown in
Fig.~\ref{tab:annotationexample}.  The annotations provide a domain
language for an open-ended set of emotion description
questions.  Some questions, like ``is it positive?'' are seen many
times, while some are rare or only occur once.  Similary, many
emotions, like ``happy'' and ``sad'' were seen many times while other
emotions like ``maudlin'' were seen only once.  The questions together
with the answers (yes, no, or other type of answers including
hedges \cite{Kazemzadeh2013}) form descriptions of emotions.  The
annotations we describe are ways to represent these questions from the
EMO20Q game.

The annotations of EMO20Q are based on a logical way of representing
questions about emotions.  The emotion being asked about in the EMO20Q
game is represented as a variable $e$, and the questions are
represented as predicates involving that emotion.  The question is
represented as a declarative statement which the answerer can agree to
by answering 'yes' or disagree with by answering 'no'.  In practice,
answers may contain hedges and answers that span between yes and
no. Since the annotations are a type of predicate logic, there could
be different ways of expressing the same proposition, so there could
be ambiguity.  This ambiguity poses a problem in how the annotations
are currently used.  The actual predicate logic statements are not
decomposed, parsed, and reasoned about, but instead they are treated
as a single, unitary propositional label for the question, which is
then used as a feature by the sequential Bayes ML model used by the
dialog agent.  Thus, the feature vector used by the dialog agent
represents each emotion word as a set of (proposition, answer) pairs,
where the proposition is the string value of the predicate label.  For
a more detailed description of the data collection and annotation
process see \cite[Section~6.2.3]{Kazemzadeh2013a}.

The annotation was done by a single annotator, so it is likely that
there is some errors or inconsistencies.  The twenty question game
format and the sequential Bayes algorithm we used made the agent
tolerant to annotation mistakes.  In fact, tolerance of errors in
agents is a design principle inspired by human communication:
sequential human-human interactions in dialog format has many error
correcting behaviors.

Although the existing annotations successfully allowed for an EMO20Q
question asking agent to be created and trained on new human-computer
dialogs \cite{Kazemzadeh2016}, the annotations were not explicitly
assessed. The annotations were created by a single annotator so they
have not been assessed for annotator agreement and they may have
errors or inconsistencies due to the lack of assessment.  By using the
existing annotations and observing the classification performance on a
paraphrase task, we aim to assess and correct the annotations of the
single annotator.

\subsection{Complex Annotation Tasks and Automation}
\label{sec:annotation-background}

Annotations can range from a fixed set of categorical labels that are
associated 1-to-1 with data items, to sequential labels that may have
order constraints, to complex, multifaceted structure
\cite{Mower2009,EmotionML,Burkhardt2014,Davani2021}.  More recently,
captioning tasks involve associating unstructured descriptions as
annotations of data.  This work fits somewhat between a complex,
structured annotation and a free form description.  The annotation has
a logical structure like a domain language.  However, the syntax is
not checked and the phenomenon that the annotation seeks to describe,
dialog questions about emotions, can be ambiguous and have tricky edge
cases. There should be a single annotation per user turn, as opposed
to the captioning task annotations, where one image may have multiple
labels.

Since annotation is a bottleneck for ML tasks, it has been the focus
of process optimization \cite{Hovy2010,Perry2021}, especially
automation. Automation approaches range from \emph{pre-annotation},
i.e. annotating the data with automation first and then having
annotators verify it \cite{Montani2021} often using \emph{transfer
learning} to apply models from another task to explore a new corpus
\cite{Felt2014}, \emph{incremental learning} incrementally adjusting
models as more annotations are observed \cite{Schulz2019}, to \emph{
active learning}, where automated processes select the most
informative data instances to present to annotators
\cite{Zhu2010,Hovy2012}.  These cases focus on speeding up the
annotation process but they do not inherently assess annotator
accuracy because this is usually determined by having multiple
annotators.  In our case, the annotation assistance is after-the-fact
and we focus on detecting annotator errors. 

Another trend aspect of annotation research is a spectrum between
controlled environments like pervasive computing and
internet-of-things \cite{Woznowski2017} to more controlled
environments like medical image annotation \cite{Gobbel2014}.  In less
controlled environments, annotators may not be trained, whereas in
more controlled environments, like medical domains, the annotators are
usually highly trained.  In the case of the EMO20Q data, the
human-human games are a minimally controlled observational study
(natural experiment), but the human-computer games have more ability
to implement experimental controls.  The annotation task itself in EMO20Q
theoretically motivated, but because there is only one annotator, it
is similar to less controlled annotation tasks.







\section{Methodology}
\label{sec:methodology}

The paraphrase task in ML seeks to predict whether two sentences are
paraphrases of each other or not.  Abstractly speaking, it is a binary
classification task with two sentences as inputs.  If the two
sentences are paraphrases, then the prediction is $1$, else $0$. First
we will describe the creation of the training set from the annotated
dialog data.  Then we will describe the mapping of the training set
into the format that BERT requires and the fine-tuning of this model.

\subsection{Data}

The annotated human-human dialog data was converted into the input for
the ML paraphrase task using the intuition that sentences with the
same annotation should be paraphrases of each other.  By carefully
following an annotation standard, we could be sure that string
equality of annotations would test whether two questions in
the dialog were paraphrases or not.

There were 110 EMO20Q dialog games from 25 participants with a total
of 1223 annotated question turns, with 700 unique annotations.  Most
(553 of 700) of the unique annotations were hapaxes (i.e. annotations
with a single occurence).  These hapaxes were excluded from the ML
training because it would not be possible to evalutate them.  With the
hapaxes excluded, there were 543 questions with 147 unique
annotations. Data and code for EMO20Q are available on
Github~\cite{emo20q-github}.

Although the data left over after removing the hapaxes seems to be
quite small for NLP tasks, the creation of input to the paraphrase
tasks takes advantage of the combinatorial expansion of the data into
pairs.  Because the input to the paraphrase task is sentence pairs and
the order of sentences in pairs will result in unique inputs, the
resulting number of data instances for the paraphrase task is an n-choose-k permutation:
\begin{eqnarray*}
  P(n,k) = \frac{n!}{(n-k)!} = \frac{543!}{(543-2)!} = 543*542 = 294306
\end{eqnarray*} which is actually more data than the Microsoft Research Paraphrases
(MSRP) dataset \cite{Dolan2005}.  There are many more negative
examples (sentence pairs that are not paraphrases) than positive ones:
the 4588 positive examples account for only approximately 1.5\% of the
total paraphrase instances. If we had tried to balance positive and
negative instances by downsampling the negative instances, then the
size of the data would have been comparable to the MSRP. As we show in
Section~\ref{sec:results}, the imbalance of positive and negative
training samples does not adversely affect performance.

These 294306 paraphrase instances were randomly split into training,
validation, and test set partitions: 200,000 (\texttildelow 68\%) were
in the training set, 14,306 (\texttildelow 5\%) in the validation set,
and 80,000 (\texttildelow 27\%) in the test set. Each partition had
approximately the same ratio of positive and negative examples.

\subsection{Model}

The BERT model was fine-tuned from the training data using the
Tensorflow library~\cite{Abadi2016} and official models, version
2.4.0~\cite{Tensorflow2021}. The specific model used was the uncased
English model trained on Wikipedia and the BookCorpus, with 12 hidden
layers of size 768, with 12 attention heads
\cite{Devlin2019,bert-uncased-tensorflowhub}.

The text data was tokenized using the default BERT tokenizer with
vocabulary size 30522.  The sentence pairs were encoded into inputs by
concatenating the class token (``[CLS]''), sentence~1's tokens, the
separator token (``[SEP]''), and then sentence~2's tokens.

The BERT model was then fine-tuned with default configurations in
batch sizes of 32 for 2 epochs using the Adam optimizer, with 6250
steps per epoch, a learning rate of $2e-5$, and 1250 warmup
steps. Each epoch took approximately 25 minutes on a Colab notebook
with a GPU backend.\footnote{For more details, the code used to
replicate this experiment can be found at
https://colab.research.google.com/drive/1sXqTT2DoJ1hFBR02iM8W4GKZe5OxCYJy
}

\section{Results}
\label{sec:results}


The models trained using the described methodology achieved
$99.9725$\% accuracy.  Because the negative instances greatly
outnumber the positive examples, the majority baseline classifier
would achieve an already high accuracy of $98.44$. In terms of error,
the model's error of $0.0275$\% represents less than one fiftieth of
the majority class baseline error of $1.56$\%. The model achieves
$98.46$\% recall, $99.65$\% precision, and F1 score of $99.03$\%.

It is important to remember that these results represent unverified
annotations.  Thus, some of the errors are not true errors due to
misclassifications, but rather annotation errors. Despite having
80,000 test instances, the high accuracy resulted in only 22 instances
where the predicted labels differed from the annotated labels.
Therefore, it was feasible for to manually examine each case where the
predicted label differed from the annotated
label.

Table~\ref{tab:error-analysis} gives a listing of the 22 errors and a
classification of whether they were true model prediction errors
(\emph{pred.}), annotation errors (\emph{ann.}), or data preparation
errors (\emph{prep.}). For example, in error \#1 in the table,
hypothetically the model failed to learn that the phrase ``Do you feel
like this when someone close to you dies?'' is a paraphrase of ``do
you feel it when someone dear has passed away?'', so this would be a
model prediction error.  Similarly in example \#2, the model failed to
understand the user's spelling error, ``jealosy''.  This is asking a
lot of a model to understand spelling mistakes, but users do make
typing mistakes and a human would still understand the variation in
spelling, so we consider this a model error as well.

In example \#3 of Table~\ref{tab:error-analysis}, the two sentences
have different annotations (``associated(e,disappointment)''
vs. ``similar(e,disappointment)''), so it was not considered a
paraphrase based on annotations.  However, the model predicted it to
be a paraphrase and it is in fact indicative of ambiguity in the
annotations and associated annotator variation, so we consider it an
annotation error.  Similarly, examples \#13 and \#19 show errors that
are due to a similar annotation errors (``similar(e,depression)''
vs. ``associated(e,depression)'' and ``associated(e,otherPeople)''
vs. ``associated(e,otherPerson)'').

Examples \#4 and \#22 Table~\ref{tab:error-analysis} show data
preparation errors.  In these cases, a script that was used to
anonymize the data by removing user names accidentally replaced
instances of the string ``test'' because there was a user that chose
that is his/her username.  This shows that the use of performant ML
models can be used to find other issues in data preparation besides
annotation.

\begin{table*}
  \caption{Error analysis: prediction errors (pred.) vs. annotation errors (ann.) vs. data ppreparation errors (prep.).}
  \label{tab:error-analysis}
  \centering
  \small
  \renewcommand{\arraystretch}{.3}
  \begin{tabular}{| c |l | c | c | c |} 
    \hline
    \#
    &\makecell[l]{Sentence 1 \\ Sentence 2} 
    & Annotated
    & Predicted
    & \makecell[l]{Error \\ Type} 
    \\
    \hline

    1
    &\makecell[l]{
      \scriptsize Do you feel like this when someone close to you dies? \\
      \scriptsize do you feel it when someone dear has passed away?} 
    & 1
    & 0
    & pred.
    \\
    \hline

    2
    &\makecell[l]{
      \scriptsize jealosy? \\
      \scriptsize haha, jealousy?} 
    & 1
    & 0
    & pred.
    \\
    \hline
    
    3
    &\makecell[l]{
      \scriptsize id it related to sth disappointing? *is \\
      \scriptsize similar to disappointed?} 
    & 0
    & 1
    & ann.
    \\
    \hline
    
    4
    &\makecell[l]{
      \scriptsize would you feel it if you had an exam the next day?\\
      \scriptsize would you feel it if you had a user13 the next day?} 
    & 1
    & 0
    & prep.
    \\
    \hline
    
    5
    &\makecell[l]{
      \scriptsize stronger thanoverwhelmed? \\
      \scriptsize is it more intense than overwhelmed?} 
    & 1
    & 0
    & pred.
    \\
    \hline
    
    6
    &\makecell[l]{
      \scriptsize do you feel it when someone dear has passed away? \\
      \scriptsize would you feel it if someone close to you had died? } 
    & 1
    & 0
    & pred.
    \\
    \hline
    
    7
    &\makecell[l]{
      \scriptsize is it melancholic? \\
      \scriptsize is it less severe that depressed, sth like melancholic} 
    & 1
    & 0
    & pred.
    \\
    \hline
    
    8
    &\makecell[l]{
      \scriptsize is it like misery? \\
      \scriptsize misery? } 
    & 1
    & 0
    & pred.
    \\
    \hline
    
    9
    &\makecell[l]{
      \scriptsize is it like being optimistic? \\
      \scriptsize is it kinda like being optmistic?} 
    & 1
    & 0
    & pred.
    \\
    \hline
    
    10
    &\makecell[l]{
      \scriptsize is it kinda like being optmistic? \\
      \scriptsize is it like being optimistic?} 
    & 1
    & 0
    & pred.
    \\
    \hline
    
    11
    &\makecell[l]{
      \scriptsize id it related to sth disappointing? *is \\
      \scriptsize is it close to being disappointed} 
    & 0
    & 1
    & ann.
    \\
    \hline
    
    12
    &\makecell[l]{
      \scriptsize Jealousy? \\
      \scriptsize jealosy?} 
    & 1
    & 0
    & pred.
    \\
    \hline
    
    13
    &\makecell[l]{
      \scriptsize is it like depression? \\
      \scriptsize so would this be an emotion that might be assimilated to depression} 
    & 0
    & 1
    & ann.
    \\
    \hline
    
    14
    &\makecell[l]{\scriptsize is it shy? \\ is it like being shy?} 
    & 1
    & 0
    & pred.
    \\
    \hline
    
    15
    &\makecell[l]{
      \scriptsize do you feel it when someone dear has passed away? \\
      \scriptsize Do you feel like this when someone close to you dies?} 
    & 1
    & 0
    & pred.
    \\
    \hline
    
    16
    &\makecell[l]{
      \scriptsize so more like plain happy? \\
      \scriptsize is it similar to happy?} 
    & 1
    & 0
    & pred.
    \\
    \hline
    
    17
    &\makecell[l]{
      \scriptsize is it associated with being aggravated? \\
      \scriptsize how about with aggravation?
    } 
    & 1
    & 0
    & pred.
    \\
    \hline
    
    18
    &\makecell[l]{
      \scriptsize would you most likely feel this towards someone you don't know? \\
      \scriptsize thanks, do you feel the emotion towards strangers?} 
    & 1
    & 0
    & pred.
    \\
    \hline
    
    19
    &\makecell[l]{
      \scriptsize is there another person involved? \\
      \scriptsize does it relate to how you feel about other people?} 
    & 0
    & 1
    & ann.
    \\
    \hline
    
    20
    &\makecell[l]{\scriptsize does it relate to how you feel about other people? \\
      is this emotion always related to another persons influence? } 
    & 1
    & 0
    & pred.
    \\
    \hline
    
    21
    &\makecell[l]{
      \scriptsize felt during betrayal?\\
      \scriptsize are there other situations when you'd feel this emotions besides betrayal?} 
    & 0
    & 1
    & pred.
    \\
    \hline
    
    22
    &\makecell[l]{
      \scriptsize would you feel it if you had a user13 the next day? \\
      \scriptsize would you feel it if you had an exam the next day?} 
    & 1
    & 0
    & prep.
    \\
    \hline

  \end{tabular}
\end{table*}

\section{Discussion}
\label{sec:discussion}

This paper demonstrated using the BERT neural LM model to provide
feedback to a complex semantic annotation task by identifying
annotation errors.
This work specifically used the paraphrase task and data from the
EMO20Q project.  We speculate that this method could be used in other
labelling tasks, including simpler, lower-cardinality label sets. The
paraphrase task could also allow enable  non-labelling
annotation tasks, such as same vs. different meaning annotations.

As described in Section~\ref{sec:annotation-background}, there are a
number of other annotation automation techniques.  Using this approach together
with others could help to further optimize the annotation
process. While the other annotation automation techniques we saw
involve pre-annotation or active learning, this work focused on
after-the-fact assessment, so the approaches could be
complimentary. Integration of this work with active
learning, pre-annotation, and determining how much data is sufficient
is an open question we hope to address in future work.

One limitation is that this approach left out hapax annotation labels
and required at least two instances of each annotation label.  One way
to deal with this could be to use the neural networks to generate
artifical data instances to show the user.  In this case, the hapax
labels could be used to generate new, artifical paraphrases and verify
whether the user would use the same label for the artifical instance.

This approach can assist the annotator by showing potential annotation
errors.  However, it still requires human intervention to fix the
annotations.  Automating the correction or providing suggestions
(e.g. merge two labels into one or separate one label into two or
more) is another direction for future work. There has been recent work
in dealing with bias in annotation \cite{Davani2021}.  Having an
automated assistant to ensure consistent annotations could be a way to
avoid bias.

Finally, this method was made possible by the high performance of
modern ML models like BERT.
It is an open
question as to further optimization of the default BERT settings.
Also it is an open question about using
traditional ML models in a similar approach.
A similar approach using less performant models could be achieved by
looking at performance errors on the training or validation sets.






\section{Conclusion}
\label{sec:conclusion}

This paper demonstrates that the high performance of the BERT neural
ML models makes it possible for the model to give feedback to a
complex semantic annotation task.  This feedback made it feasible
to perform annotation by a single annotator and use the
machine-learned model to find annotation errors.

We acknowledge the University of St. Thomas Graduate Programs in
Software's Center for Applied 
Artificial Intelligence,
the University
of Minnesota Computational Linguistics Reading Group organizers Borui
(Bri) Zhang and Brian Reese, and ARDUOUS 2022 reviewers for their
support and feedback.




\begin{thebibliography}{10}

\bibitem{Mower2009}
E.~Mower, A.~Metallinou, C.-C. Lee, A.~Kazemzadeh, C.~Busso, S.~Lee, and
  S.~Narayanan, ``Interpreting ambiguous emotional expressions,'' in {\em ACII
  Special Session: Recognition of Non-Prototypical Emotion from Speech- The
  Final Frontier?}, (Amsterdam), pp.~1 -- 8, IEEE, 2009.

\bibitem{Davani2021}
A.~M. Davani, M.~Díaz, and V.~Prabhakaran, ``Dealing with disagreements:
  Looking beyond the majority vote in subjective annotations.'' arXiv, 2021.

\bibitem{Firestone2020}
C.~Firestone, ``Performance vs. competence in human–machine comparisons,''
  {\em Proceedings of the National Academy of Sciences (PNAS)}, vol.~117,
  pp.~26562--26571.

\bibitem{Eckersley2017}
P.~Eckersley and Y.~Nasser, ``{EFF} {AI} progress measurement project (2017-),
  accessed 2021-11-19.''
\newblock Available at \url{https://eff.org/ai/metrics}.

\bibitem{Vaswani2017}
A.~Vaswani, N.~Shazeer, N.~Parmar, J.~Uszkoreit, L.~Jones, A.~N. Gomez,
  L.~Kaiser, and I.~Polosukhin, ``Attention is all you need,'' {\em Proceedings
  of the 31st International Conference on Neural Information Processing Systems
  (NIPS)}, 2017.

\bibitem{Manning2020}
C.~D. Manning, K.~Clark, J.~Hewitt, U.~Khandelwal, and O.~Levy, ``Emergent
  linguistic structure in artificial neural networks trained by
  self-supervision,'' {\em Proceedings of the National Academy of Sciences
  (PNAS)}, 2020.

\bibitem{Turc2019}
I.~Turc, M.-W. Chang, K.~Lee, and K.~Toutanova, ``Well-read students learn
  better: On the importance of pre-training compact models,'' {\em arXive},
  2019.

\bibitem{Kazemzadeh2013a}
A.~Kazemzadeh, {\em Natural Language Description of Emotion}.
\newblock PhD thesis, University of Southern California, 2013.

\bibitem{Kazemzadeh2013}
A.~Kazemzadeh, S.~Lee, and S.~Narayanan, ``Fuzzy logic models for the meaning
  of emotion words,'' {\em IEEE Computational Intelligence Magazine}, 2013.

\bibitem{Kazemzadeh2016}
A.~Kazemzadeh, J.~Gibson, P.~G. Georgiou, S.~Lee, and S.~S. Narayanan, ``A
  socratic epistemology for verbal emotional intelligence,'' {\em PeerJ
  Computer Science}, vol.~2, p.~e40.

\bibitem{EmotionML}
M.~Schr\"oder, P.~Baggia, F.~Burkhardt, C.~Pelachaud, C.~Peter, and E.~Zovato,
  ``{W3C} candidate recommendation: Emotion markup language ({EmotionML})
  1.0,'' May 2012.
\newblock http://www.w3.org/TR/emotionml/.

\bibitem{Burkhardt2014}
F.~Burkhardt, C.~Becker-Asano, E.~Begoli, R.~Cowie, G.~Fobe, P.~Gebhard,
  A.~Kazemzadeh, I.~Steiner, and T.~Llewellyn, ``Application of emotionml,'' in
  {\em Emotion, Social Signals, Sentiment, and Linked Open Data (ES3LOD) 2014},
  (Reykjavik), pp.~1--5, European Language Resources Association, 2014.

\bibitem{Hovy2010}
E.~Hovy and J.~Lavid, ``Towards a ‘science’ of corpus annotation: A new
  methodological challenge for corpus linguistics,'' {\em International Journal
  of Translation}, vol.~22, no.~1, 2010.

\bibitem{Perry2021}
T.~Perry, ``Lighttag: Text annotation platform,'' Sept. 2021.

\bibitem{Montani2021}
I.~Montani and M.~Honnibal, ``Prodigy {A}nnotation {T}ool,'' 2021.

\bibitem{Felt2014}
P.~Felt, E.~Ringger, K.~Seppi, and K.~Heal, ``Using transfer learning to assist
  exploratory corpus annotation,'' in {\em Proceedings of the Ninth
  International Conference on Language Resources and Evaluation ({LREC}'14)},
  (Reykjavik, Iceland), pp.~140--145, European Language Resources Association
  (ELRA), May 2014.

\bibitem{Schulz2019}
C.~Schulz, C.~M. Meyer, J.~Kiesewetter, M.~Sailer, E.~Bauer, M.~R. Fischer,
  F.~Fischer, and I.~Gurevych, ``Analysis of automatic annotation suggestions
  for hard discourse-level tasks in expert domains,'' in {\em Proceedings of
  the 57th Annual Meeting of the Association for Computational Linguistics},
  (Florence, Italy), pp.~2761--2772, Association for Computational Linguistics,
  July 2019.

\bibitem{Zhu2010}
J.~Zhu, H.~Wang, E.~Hovy, and M.~Ma, ``Confidence-based stopping criteria for
  active learning for data annotation,'' {\em ACM Transactions on Speech and
  Language Processing (TSLP)}, vol.~6, no.~3, pp.~1--24, 2010.

\bibitem{Hovy2012}
D.~Hovy and E.~Hovy, ``Exploiting partial annotations with {EM} training,'' in
  {\em Proceedings of the {NAACL}-{HLT} Workshop on the Induction of Linguistic
  Structure}, (Montr{\'e}al, Canada), pp.~31--38, Association for Computational
  Linguistics, June 2012.

\bibitem{Woznowski2017}
P.~Woznowski, E.~Tonkin, P.~Laskowski, N.~Twomey, K.~Yordanova, and A.~Burrows,
  ``Talk, text or tag?,'' in {\em PerCom Workshop Proceedings (1st
  International Workshop on Annotation of useR Data for UbiquitOUs Systems
  (ARDUOUS))}, 2017.

\bibitem{Gobbel2014}
G.~T. Gobbel, J.~Garvin, R.~Reeves, R.~M. Cronin, J.~Heavirland, J.~Williams,
  A.~Weaver, S.~Jayaramaraja, D.~Giuse, T.~Speroff, S.~H. Brown, H.~Xu, and
  M.~E. Matheny, ``{Assisted annotation of medical free text using RapTAT},''
  {\em Journal of the American Medical Informatics Association}, vol.~21,
  pp.~833--841, 01 2014.

\bibitem{emo20q-github}
``Emotion twenty questions github respository (accessed 2021-11-19).''
\newblock https://github.com/abecode/emo20q.

\bibitem{Dolan2005}
W.~B. Dolan and C.~Brockett, ``{Automatically Constructing a Corpus of
  Sentential Paraphrases},'' in {\em Proceedings of the Third International
  Workshop on Paraphrasing ({IWP}2005)}, 2005.

\bibitem{Abadi2016}
M.~Abadi, P.~Barham, J.~Chen, Z.~Chen, A.~Davis, J.~Dean, M.~Devin,
  S.~Ghemawat, G.~Irving, M.~Isard, M.~Kudlur, J.~Levenberg, R.~Monga,
  S.~Moore, D.~G. Murray, B.~Steiner, P.~Tucker, V.~Vasudevan, P.~Warden,
  M.~Wicke, Y.~Yu, and X.~Zheng, ``Tensorflow: A system for large-scale machine
  learning,'' {\em 12th USENIX Symposium on Operating Systems Designand
  Implementation (OSDI ’16)}, 2016.

\bibitem{Tensorflow2021}
``Tensorflow model garden: Models and examples built with tensorflow (accessed
  2021-11-19).''
\newblock https://github.com/tensorflow/models.

\bibitem{Devlin2019}
J.~Devlin, M.-W. Chang, K.~Lee, and K.~Toutanova, ``{BERT}: {P}re-training of
  {D}eep {B}idirectional {T}ransformers for {L}anguage {U}nderstanding,'' in
  {\em Proceedings of the 2019 Conference of the North {A}merican Chapter of
  the Association for Computational Linguistics: Human Language Technologies},
  NAACL, NAACL, 2019.

\bibitem{bert-uncased-tensorflowhub}
``Tensorflow hub: Bert uncased l-12 h-768 a-12 (accessed 2021-11-19).''
\newblock https://tfhub.dev/google/bert\_uncased\_L-12\_H-768\_A-12/1.

\end{thebibliography}

\end{document}